\documentclass{article}

\usepackage[preprint]{corl_2025} 

\usepackage{multicol}
\usepackage{graphicx}
\usepackage{xspace}
\usepackage{xcolor}
\usepackage{caption}

\usepackage{pifont}

\usepackage{wrapfig}
\usepackage{bbding}  
\usepackage{pifont}  

\usepackage{booktabs}
\usepackage{float}
\usepackage{amsmath}  
\usepackage{amssymb}  
\usepackage{multirow, booktabs}
\usepackage{siunitx}

\newcommand{\oursfull}[0]{Teleoperated Whole-Body Imitation System (TWIST)\xspace}
\newcommand{\ours}[0]{TWIST\xspace}

\newcommand{\ourwebsite}[0]{\href{https://humanoid-teleop.github.io
}{humanoid-teleop.github.io
}\xspace}

\usepackage{colortbl}
\definecolor{ourcolor}{HTML}{99e0eb}
\definecolor{ourblue}{HTML}{27a2c3}

\definecolor{tablecolor}{HTML}{ccf2f5} 

\definecolor{tablecolor2}{HTML}{ffcdb4}
\definecolor{citecolor}{HTML}{fe7b5b}
\definecolor{grey}{rgb}{0.9, 0.9, 0.9}
\usepackage{amssymb}

\usepackage{listings}
\definecolor{gred}{rgb}{0.859,0.267,0.216}
\definecolor{ggreen}{rgb}{0.059,0.616,0.345}

\definecolor{deepblue}{HTML}{1e7fa0}

\definecolor{deepred}{HTML}{7c2320}

\usepackage{mathrsfs}
\usepackage{amssymb}
\usepackage{amsmath} 
\usepackage{algorithmic}
\usepackage{graphicx}
\usepackage{textcomp}
\usepackage{xcolor}

\usepackage[font=small]{caption}
\usepackage[compact]{titlesec}

\definecolor{deepgreen}{RGB}{63, 126, 49}
\definecolor{deepred2}{RGB}{196, 49, 25}

\usepackage{enumitem}

\setlist[itemize]{leftmargin=1em}

\title{

TWIST: Teleoperated Whole-Body Imitation System

}

\author{
Yanjie Ze$^{1*}$\quad Zixuan Chen$^{2*}$\quad João Pedro Araújo$^{1*}$\quad Zi-ang Cao$^{1}$\\
\textbf{Xue Bin Peng$^{2}$}\quad \textbf{Jiajun Wu}$^{1\dag{}}$\quad \textbf{C. Karen Liu}$^{1\dag{}}$\vspace{0.05in}\\
 $^1$Stanford University $^2$Simon Fraser University\vspace{0.05in} $^*$Equal Contribution $^{\dag{}}$Equal Advising\\
 \href{https://humanoid-teleop.github.io}{\textsc{\textbf{\color{deepred}humanoid-teleop.github.io}}}
 }

\begin{document}
\maketitle

\begin{figure}[htbp]
\vspace{-0.4in}
    \centering
\includegraphics[width=0.8\linewidth]{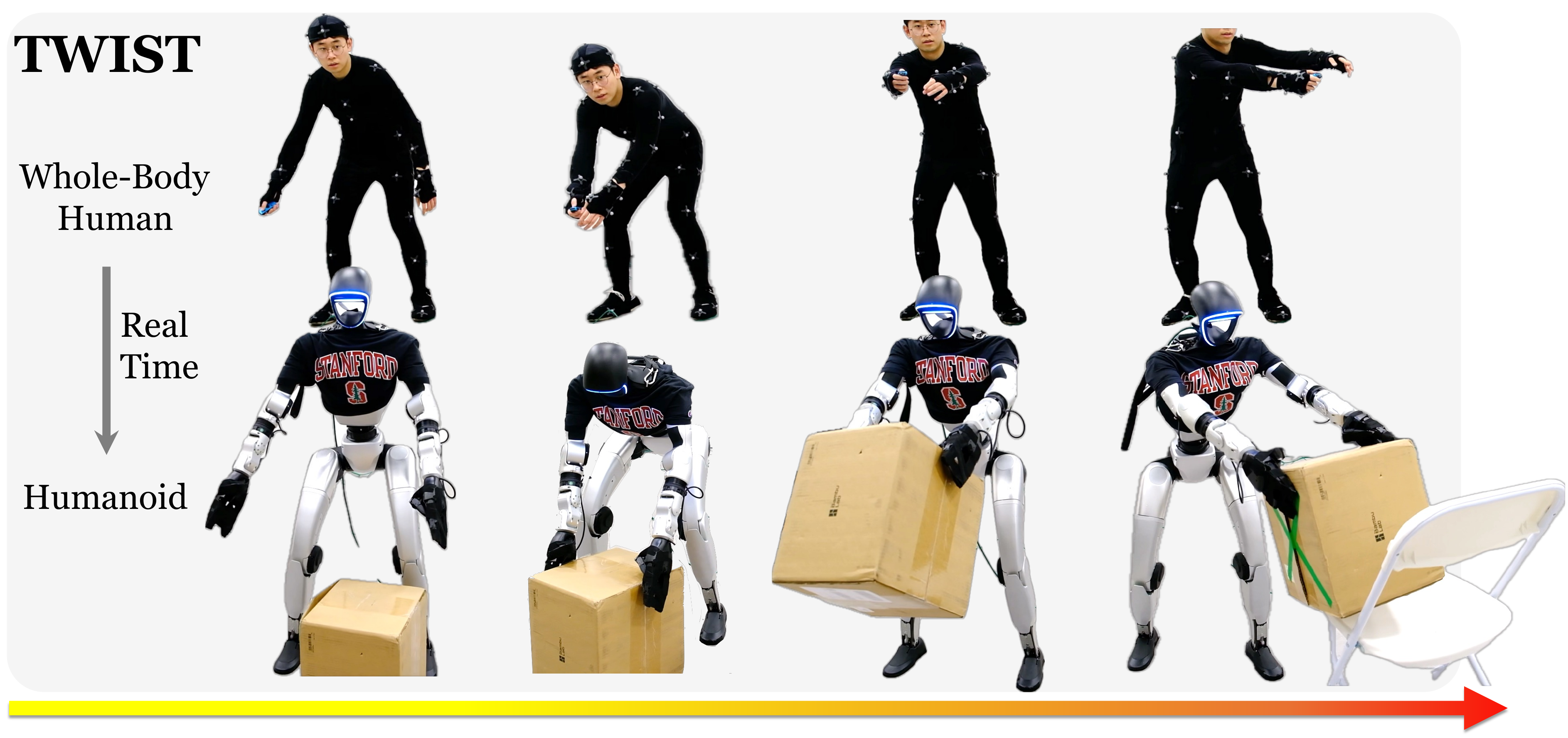}
\vspace{-0.05in}
    \caption{The \textbf{\oursfull} is a system that teleoperates humanoid robots with real-time whole-body human data and a \textit{single} neural network controller. \ours achieves versatile, coordinated, whole-body skills that are not present in previous works. }
    \label{fig:teaser}
    \vspace{-0.2in}
\end{figure}

\begin{abstract}
Teleoperating humanoid robots in a whole-body manner marks a fundamental step toward developing general-purpose robotic intelligence, with human motion providing an ideal interface for controlling all degrees of freedom. Yet, most current humanoid teleoperation systems fall short of enabling coordinated whole-body behavior, typically limiting themselves to isolated locomotion or manipulation tasks. We present the Teleoperated Whole-Body Imitation System (TWIST), a system for humanoid teleoperation through whole-body motion imitation. We first generate reference motion clips by retargeting human motion capture data to the humanoid robot. 
We then develop a robust, adaptive, and responsive whole-body controller using a combination of reinforcement learning and behavior cloning (RL+BC). Through systematic analysis, we demonstrate how incorporating privileged future motion frames and real-world motion capture (MoCap) data improves tracking accuracy. TWIST enables real-world humanoid robots to achieve unprecedented, versatile, and coordinated whole-body motor skills—spanning whole-body manipulation, legged manipulation, locomotion, and expressive movement—using a \textit{single} unified neural network controller. Our website: \ourwebsite

\end{abstract}

\keywords{Humanoids, Whole-Body Teleoperation, Learning-Based Control}

\begin{figure}[htbp]
    \centering
\includegraphics[width=1.0\linewidth]{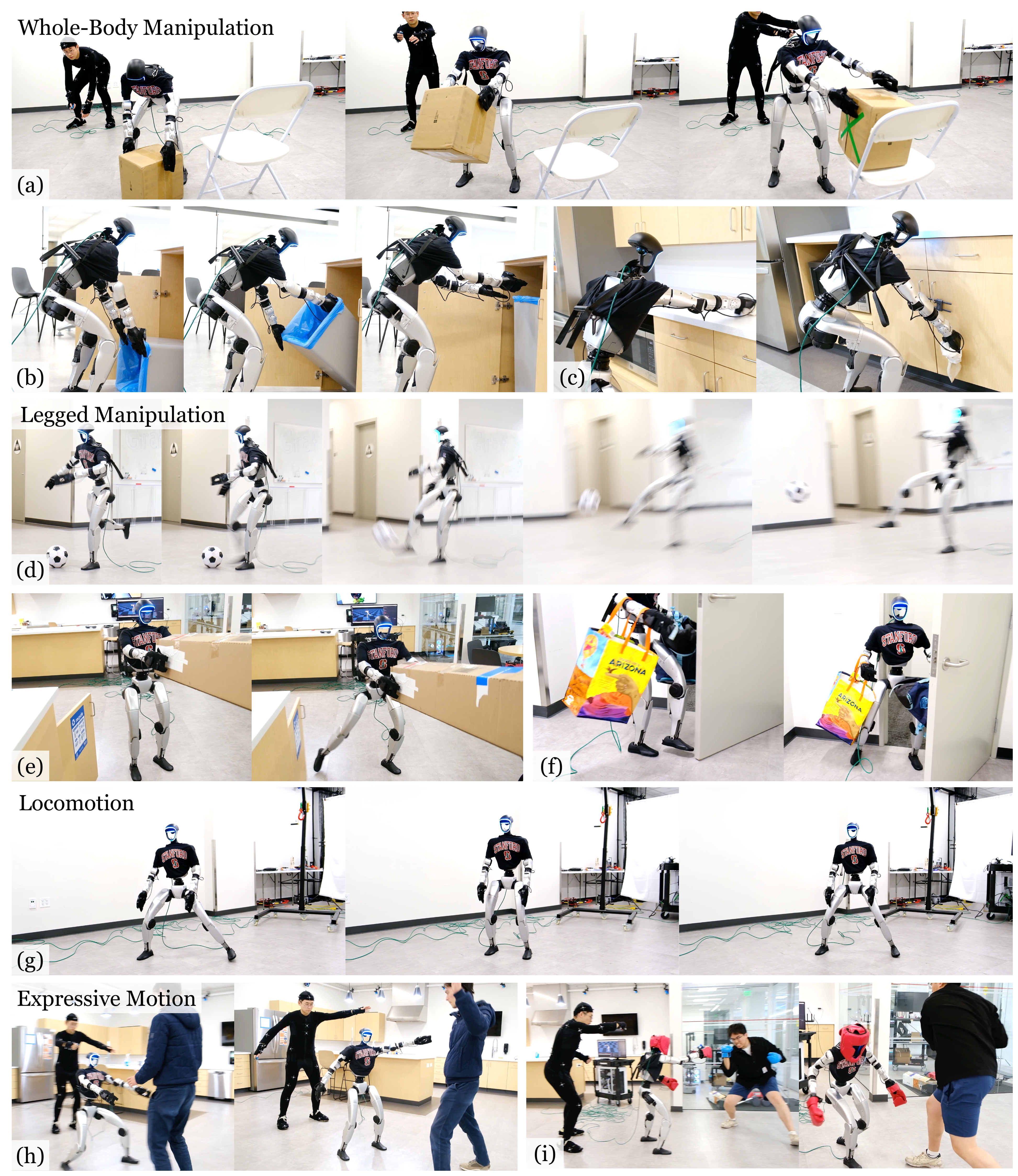}
    \caption{\textbf{The \oursfull presents versatile, coordinated, and human-like whole-body skills on real-world humanoid robots.} Our robot can perform whole-body manipulation (\textit{e.g.}, lifting boxes from the ground), legged manipulation (\textit{e.g.}, kicking the football), locomotion (\textit{e.g.}, walking sideways), and expressive motions (\textit{e.g.}, Waltz dance). More videos: \ourwebsite}
    \label{fig:task illustration}
\end{figure}

\section{Introduction}

Humans naturally master versatile, coordinated whole-body skills essential for everyday tasks.
For example, when entering a room, a person with objects in both hands can use their foot to open the door.
When cleaning, they can bend down to reach under the bed and tidy up the hidden lower space.
In sports like soccer, humans can balance on one leg while using the other to kick a ball. Enabling humanoid robots to perform such coordinated, whole-body actions is crucial for developing general-purpose robots that can live and work alongside humans in household environments.

One promising way to empower humanoid robots with versatile whole-body dexterity is by imitating human movements. However, due to the embodiment gap between humanoids and humans, simply imitating offline human motion data is insufficient for humanoid visuomotor control; we must build a whole-body teleoperation system to acquire humanoid observation-action data.  
Building a capable whole-body teleoperation system, however, has long been a challenge~\citep{penco2019multimode, dafarra2024icub3, penco2018robust, darvish2023teleoperation, darvish2019whole, he2024learning, fu2024humanplus, he2024omnih2o, he2024hover}. Classical methods~\citep{koenemann2014real, penco2019multimode, dafarra2024icub3, penco2018robust, darvish2023teleoperation, darvish2019whole} use modular model-based controllers to separately handle teleoperation and balance,
limiting the system’s whole-body capabilities and robustness. Recent learning-based controllers have shown great promise~\citep{he2024learning, fu2024humanplus, he2024omnih2o, he2024hover}, but the coordinated whole-body skills, \textit{e.g.}, crouching down to lift a box on the ground, remain limited,  mainly due to the lack of accurate, real-time whole-body tracking targets and robust controllers for tracking such diverse, real-time motions.

In this work, we propose the \textbf{\oursfull}, a humanoid teleoperation system that achieves versatile, coordinated whole-body skills by imitating whole-body human motions in real time (Figure~\ref{fig:teaser}).
A key requirement for such a system is a controller that autonomously converts arbitrary, real-time human movements into \emph{balanced} robot motions, while closely matching the human’s behaviors. To this end, we formulate whole-body teleoperation as a real-time motion retargeting and tracking problem. We first derive tracking targets—humanoid joint positions and root velocities—by retargeting arbitrary human motions captured by motion capture (MoCap) devices. We then train a \textit{single} policy with reinforcement learning (RL) in large-scale simulated environments, combined with human motion data. The resulting controller can robustly and accurately track target robot joint positions and root velocities at each timestep while maintaining whole-body balance. To address the practical challenges of real-time, whole-body motion tracking and teleoperation, our training pipeline introduces several critical techniques, including:

\begin{itemize}
    
    \item To ensure low latency of teleoperation, RL policies can only observe the reference motion for one current time step, which results in more conservative and hesitant behaviors compared to policies including future reference motions as observations. This hesitant behavior is aggravated by the real-time teleoperation system because the human demonstrator tends to compensate their own movements when experiencing the hesitant behaviors on the humanoid, leading to ineffective teleoperation control. To alleviate this issue, we propose a two-stage teacher-student framework: the teacher policy is trained with privileged access to future motion frames to learn smoother behaviors, and subsequently guides the student policy, which tracks only a single frame.

    \item Offline human motion datasets are usually high-quality and smooth, while the real-time human motions and the real-time retargeting are not that stable and smooth, causing a distribution shift for online teleoperation. Therefore, we collect a small-scale MoCap human dataset (150 clips) using the online MoCap and retargeting settings, combined with 15K offline motion clips as the training set for training the RL controller. Surprisingly, despite only a small set of online motions we use,  the controller performs significantly better and more stably on unseen test motions and in real-world teleoperation.

    \item During offline retargeting of human motions, we can ensure high-quality motion data through many iterations of optimization. However, for online retargeting during teleoperation, fast inference is critical, often at the expense of smoothness. We find that jointly optimizing 3D joint positions and orientations helps mitigate this offline-to-online gap, compared to optimizing orientations alone~\citep{anonymous2025mindmapping}.

    \item As the learning objective of the controller is simply motion tracking, tasks requiring force exertion (\textit{e.g.}, lifting a box) rather than reaching target positions represent out-of-distribution scenarios, causing the controller to produce jittery behaviors occasionally. To enable the controller to learn to apply force, we propose to train controllers with large end-effector perturbations, which significantly improves robustness in tasks requiring contact and force.

\end{itemize}

With all these critical components integrated, \ours achieves remarkable whole-body teleoperation capabilities on real-world humanoid robots. As shown in Figure~\ref{fig:task illustration}, \ours enables the Unitree G1—a medium-sized humanoid robot with 29 degrees of freedom (DoF)—to perform a wide range of diverse, human-like skills. All teleoperation tasks are accomplished using a \textit{single} neural network controller.

\begin{figure}[t]
    \centering
    \includegraphics[width=1.0\linewidth]{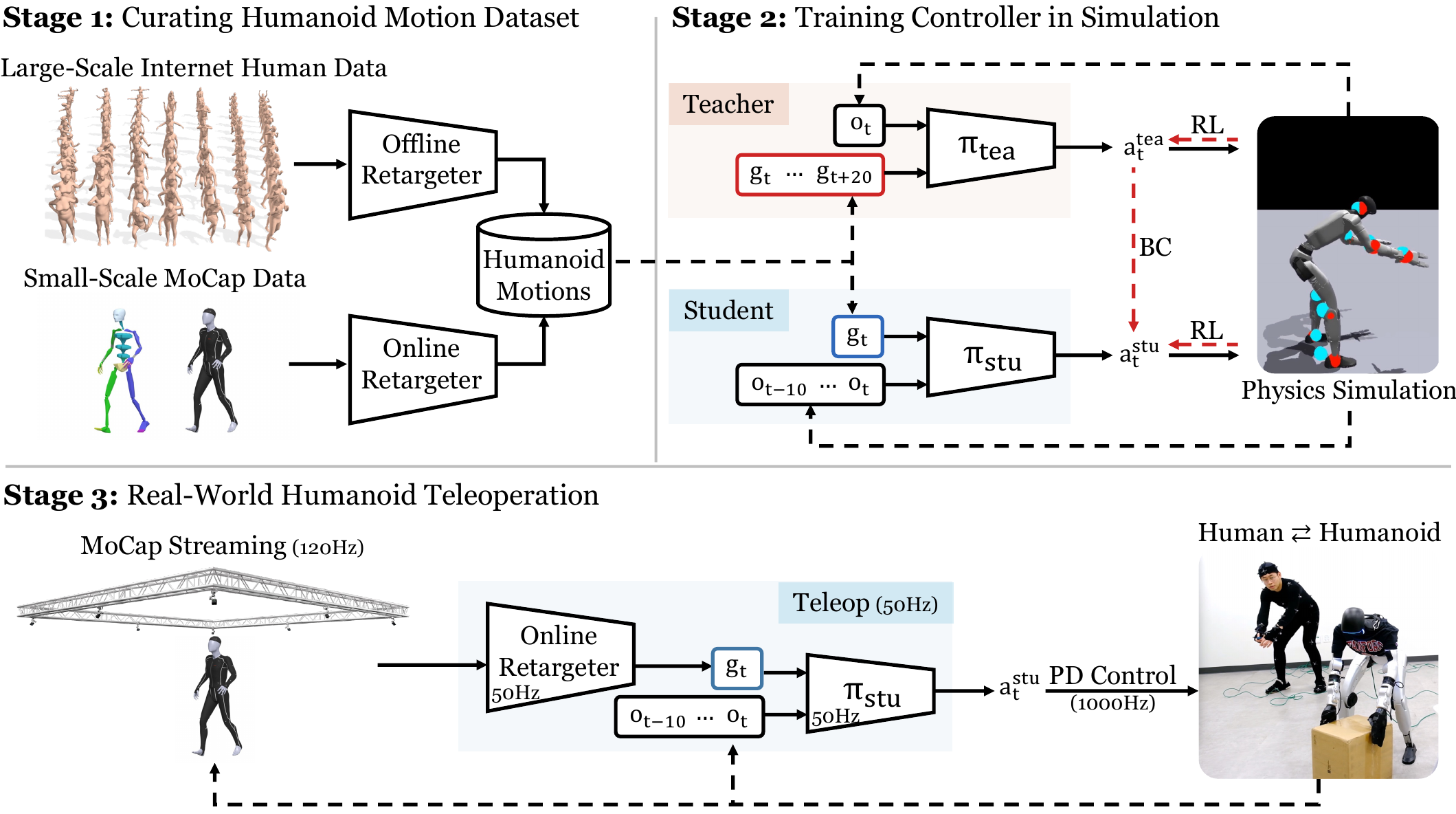}
    \vspace{-0.2in}
    \caption{\textbf{The \oursfull consists of 3 stages}: 1) curating a humanoid motion dataset by retargeting Internet human data and our MoCap data, 2) training a single whole-body controller in simulation, 3) teleoperating real-world humanoid robots with MoCap devices.
    }
    \label{fig:method}
    \vspace{-0.2in}
\end{figure}

\section{Related Works}

\textbf{Learning-Based Whole-Body Control for Humanoid Robots.}
In recent years, learning-based methods, particularly sim-to-real reinforcement learning, have made notable progress in developing whole-body controllers for humanoid robots, enabling a wide range of skills such as walking~\citep{chen2024lcp, gu2024advancing, radosavovic2024learning, radosavovic2024real, long2024hinfinity}, jumping~\citep{zhang2024wococo, fu2024humanplus}, parkour~\citep{zhuang2024humanoid_parkour}, dancing~\citep{cheng2024expressive, ji2024exbody2}, hopping~\citep{xue2025hugwbc}, and fall recovery~\citep{huang2025host, zhuang2025embrace, he2025getup}.
However, most of these works focus on developing controllers for a single specific task, limiting the generality of their methods. In contrast, our work aims to train a controller capable of performing a diverse set of real-world tasks with real-time human teleoperation.

\textbf{Teleoperation Systems for Humanoid Robots.}
Teleoperation is key to enabling humanoid robots to interact with complex real-world environments and perform manipulation tasks. Prior works have explored teleoperation modalities such as sparse VR keypoints~\citep{ze2024humanoid_manipulation, cheng2024opentv, lu2024pmp, he2024omnih2o}, exoskeletons~\citep{ben2025homie, yang2024ace, jiang2025behavior,shaw2024bimanual}, motion capture suits~\citep{dafarra2024icub3, dragan2013legibility, darvish2019whole, koenemann2014real}, and camera-based pose estimation~\citep{fu2024humanplus, he2024h2o}.
However, most systems show limited coordinated whole-body skills, making them insufficient for general household tasks. For example, Mobile-TV~\citep{lu2024pmp} and HOMIE~\citep{ben2025homie} decouple upper- and lower-body control, guiding movement with external commands like joysticks or foot pedals, but limiting whole-body tasks like kicking or obstacle traversal.
More closely related to our work, HumanPlus~\citep{fu2024humanplus} and OmniH2O~\citep{he2024omnih2o} attempt full-body teleoperation by training low-level controllers to track human motions. HumanPlus relies on camera-based pose estimation~\citep{shin2023wham}, which struggles with root position accuracy, affecting locomotion fidelity. OmniH2O uses VR keypoints that only captures upper-body movements, still lacking full whole-body control. In contrast, we introduce a whole-body teleoperation system capable of performing coordinated skills, leveraging motion capture for high-quality human data input and a robust training pipeline.

\section{Method}

We frame the problem of whole-body humanoid teleoperation as a real-time motion tracking task~\citep{fu2024humanplus,he2024omnih2o,he2024h2o,darvish2023teleoperation,peng2018deepmimic,luo2023phc}, where a whole-body controller receives real-time retargeted human motions and outputs target joint positions to make the humanoid mimic the human. This setup raises two key questions: first, how to obtain reliable real-time reference motions, and second, how to build a robust controller that can accurately track them.

\ours answers these questions through two innovations. First, we collect accurate real-time whole-body motion from a motion capture (MoCap) device; second, we develop a robust, \textit{unified} neural network controller capable of tracking diverse real-time motions through a teacher-student framework, trained with large-scale simulation and human motions. The resulting system, the \oursfull, is shown in Figure~\ref{fig:method}. In the following sections, we first describe how to curate human motion data for training the controller (Section~\ref{sec:curatemotiondata}). We then explain how to train a unified controller using this dataset (Section~\ref{sec:traincontroller}). Finally, we show how to deploy the controller for real-world teleoperation of the humanoid robot (Section~\ref{sec:realworldteleop}).

\subsection{Curating Humanoid Motion Datasets}
\label{sec:curatemotiondata}

To train a controller that presents human-like movements and accurately tracks human motions, we first curate a training dataset to serve as the fuel~\citep{peng2018deepmimic}. The majority of our data comes from publicly available MoCap datasets—AMASS~\citep{mahmood2019amass} and OMOMO~\citep{li2023omomo}—which together include over 15,000 clips (around 42 hours), with unfeasible motions such as climbing stairs being filtered out. Additionally, we collect a small in-house dataset using our own MoCap system, consisting of 150 clips (around 0.5 hours). This in-house data better reflects the conditions of real-world teleoperation, such as noise and imperfect calibration. Notably, we do not design these motions to match the teleoperation task—they are collected randomly.

Due to the embodiment gap between humanoids and humans, tracking raw human motions directly is non-trivial. Instead, we retarget them into the humanoid format to more easily compute tracking errors. For large-scale public datasets, we apply an offline retargeter similar to PHC~\citep{luo2023phc}, which optimizes key body positions. We additionally optimize temporal smoothness. While offline retargeting yields higher-quality motion, it involves iterative optimization, making it impractical for real-time teleoperation.

To simulate the real-time teleoperation setting, we use an online retargeter~\citep{anonymous2025mindmapping} on our small in-house dataset, which efficiently adjusts body orientation and foot placement using an Inverse Kinematics (IK) method~\citep{Zakka_Mink_Python_inverse_2024}. Although this approach is significantly faster and suited for real-time use, it typically results in less smooth motion, potentially impacting teleoperation performance. We mitigate this quality gap by enhancing the optimization objective of the online retargeter to jointly optimize 3D joint positions and orientations. This improvement reduces the discrepancy between offline and online motion quality, as shown in Figure~\ref{fig:bar chart} (left).

\subsection{Training A Whole-Body Controller in Simulation}
\label{sec:traincontroller}

After curating a humanoid motion dataset, we aim to train a unified whole-body controller capable of tracking arbitrary retargeted humanoid motions through large-scale simulation~\citep{makoviychuk2021isaac}. Naive approaches~\citep{fu2024humanplus} typically use single-stage reinforcement learning (RL) to train a deployable tracking policy that tracks only the current motion frame. However, this often leads to artifacts such as foot sliding and fails to produce smooth control suitable for teleoperation, largely due to the lack of access to full motion sequences. In contrast, our system adopts a two-stage approach: first, a privileged expert (teacher) policy is trained via RL with access to future reference motions. Then, a deployable student policy is jointly optimized using RL and behavior cloning (BC), relying only on proprioception and a single reference frame at each time step.

\begin{table}[t]
\centering
\vspace{-0.2in}
\begin{minipage}{0.69\textwidth}
\centering
\caption{\textbf{Reward terms and their weights.} The left table lists tracking rewards, while the middle table lists penalty terms.}
\label{tab:reward terms}
\vspace{0.5em}
\resizebox{\textwidth}{!}{%
\begin{tabular}{cc}
\begin{tabular}{ll}
\toprule
\textbf{Tracking Reward Terms} & \textbf{Weights} \\
\midrule
KeyBody Position Tracking & 2.0 \\
Joint Position Tracking & 0.6 \\
Joint Velocity Tracking & 0.2 \\
Root Pose Tracking & 0.6 \\
Root Velocity Tracking & 1.0 \\
\bottomrule
\end{tabular}
&
\begin{tabular}{ll}
\toprule
\textbf{Penalty Terms} & \textbf{Weights} \\
\midrule
Feet Contact Penalty & -5e-4 \\
Feet Slipping Penalty & -0.1 \\
Joint Velocities Penalty & -1e-4 \\
Action Rate Penalty & -0.01 \\
Feet Air Time & 5.0 \\
\bottomrule
\end{tabular}
\end{tabular}}
\end{minipage}
\hfill
\begin{minipage}{0.29\textwidth}
\centering
\caption{\textbf{Domain randomization parameters.}}
\label{tab:domain rand}
\vspace{0.5em}
\resizebox{\textwidth}{!}{%
\begin{tabular}{ll}
\toprule
\textbf{Domain Rand Params} & \textbf{Range} \\
\midrule
Base Mass (kg) & $[-3, 3]$ \\
Friction & $[0.1, 2.0]$ \\
Motor Strength & $[0.8, 1.2]$ \\
Gravity Change ($m/s 
^2$) & $[-0.1, 0.1]$ \\
Push Robot Base (m/s) & $[-0.1, 0.1]$ \\
Push End-Effector (N) & $[0,20]$ \\
\bottomrule
\end{tabular}}
\end{minipage}
\vspace{-0.2in}
\end{table}

\textbf{Privileged Teacher Policy.} The teacher policy $\pi_\text{tea}$ takes a sequence of future reference motion frames (spanning 2 seconds) as part of the input, enabling the teacher to anticipate and plan for upcoming tracking goals, further leading to smooth locomotion gaits. Besides, we choose to track joint positions and root velocities expressed in the robot’s local frame rather than the world frame,  to 1) alleviate the accumulated tracking error and imperfect retargeting~\citep{ji2024exbody2,he2024hover}; 2)  ensure better consistency with the real-world teleoperation setup. The teacher policy is optimized by PPO~\citep{schulman2017ppo,rudin2022legged_gym}, with a reward structure $r_\text{tea}$ that emphasizes accurate tracking while penalizing artifacts such as jitter and foot slippage (see Table~\ref{tab:reward terms}): $r_\text{tea} = r_\text{track} + r_\text{penalty}$.

\textbf{Deployable Student Policy.} Since privileged information (\textit{e.g.}, future motion frames) is unavailable during deployment, the expert policy $\pi_\text{tea}$ must be distilled into a deployable student policy  $\pi_\text{stu}$ using only proprioceptive inputs and immediate reference motion targets. The observational gap created by the differing input modalities poses a challenge that standard imitation methods (\textit{e.g.}, DAgger~\citep{ross2011dagger}) cannot fully address. Therefore, we adopt a hybrid RL and BC approach~\citep{radosavovic2024real}, optimizing the student policy with the following loss:
\begin{equation}
L(\pi_\text{stu}) = L_{\text{RL}}(\pi_\text{stu}) + \lambda D_{\text{KL}}(\pi_\text{stu} \parallel \pi_\text{tea}),
\label{eq:RLBC}
\end{equation}
where $L_{\text{RL}}$ denotes the PPO loss using the same reward $r_\text{tea}$ as the teacher, $D_{\text{KL}}$ denotes KL divergence for encouraging imitation of the expert, and the weight $\lambda$ is gradually reduced during training. Our results demonstrate that this RL+BC strategy significantly outperforms pure BC and pure RL, resulting in smoother motions and better generalization.

\subsection{Real-World Humanoid Teleoperation}
\label{sec:realworldteleop}

After obtaining the whole-body controller $\pi_\text{stu}$ though our two-stage training pipeline, our policy can be zero-shot deployed to the real robot, due to the careful tuning of domain randomization (see Table~\ref{tab:domain rand}). To achieve precise real-time motion tracking in the real world, we establish an online streaming and retargeting pipeline, which captures high-quality human motions at 120Hz using OptiTrack~\citep{optitrack2025} and retargets into humanoid motions at 50Hz using the retargeting approach detailed in Section~\ref{sec:curatemotiondata}. The high-quality tracking goals in the real world ensure that our robot can perform diverse tasks across manipulation and locomotion in a unified manner. Our policy $\pi_\text{stu}$ then takes the retargeted motions as input and outputs joint position commands at 50Hz on an Nvidia RTX 4090 GPU, sending to the robot's PD controller running at 1000Hz.

\begin{figure}[tb]
    \centering
    \begin{minipage}[b]{0.63\textwidth}
    \centering
    \includegraphics[width=\linewidth]{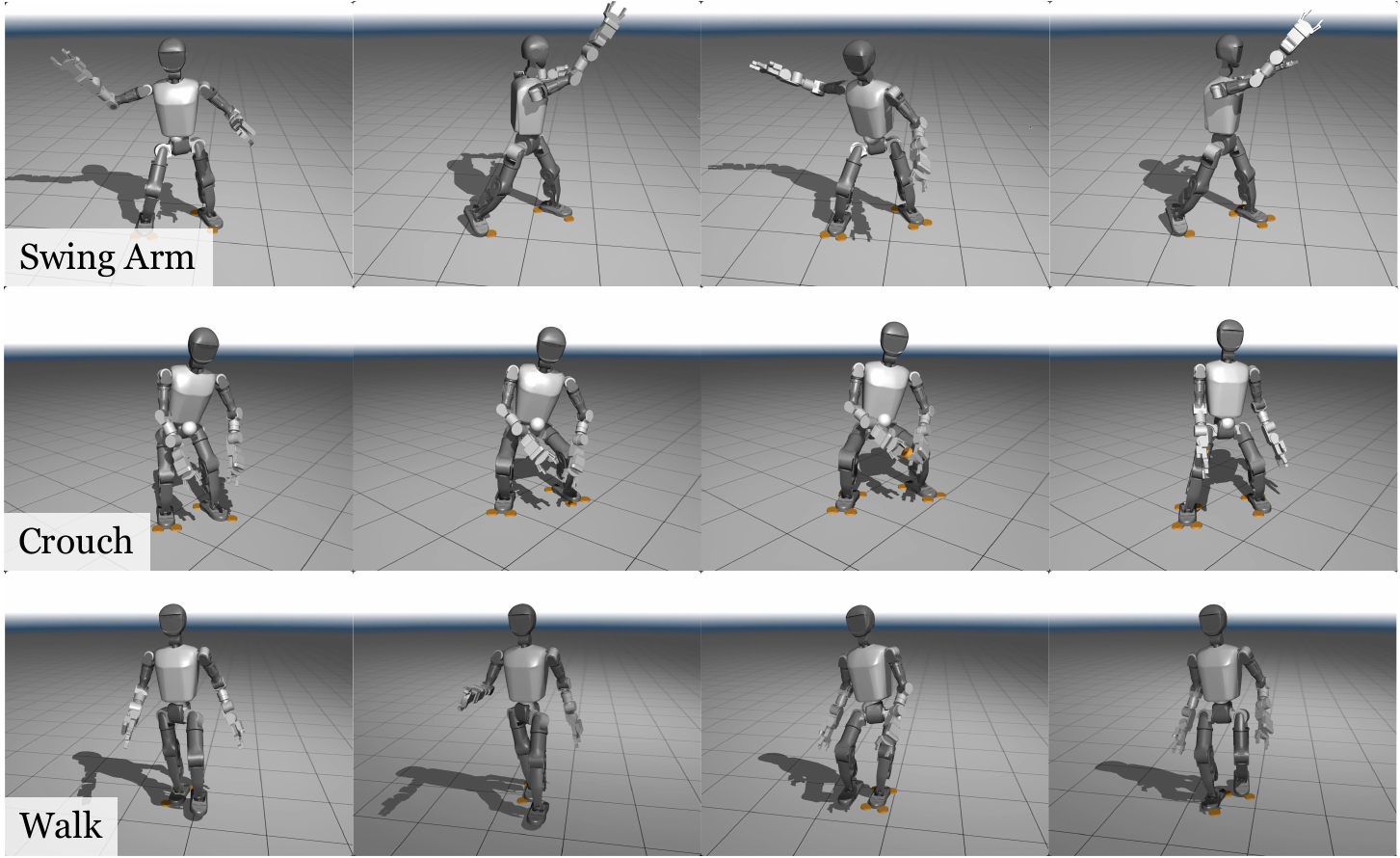}
    \caption{\textbf{Booster T1 sim2sim results.} The whole-body controller is trained in IsaacGym~\citep{makoviychuk2021isaac} and evaluated in MuJoCo~\citep{todorov2012mujoco}. The tracking goals are sampled from training data.}
    \label{fig:booster sim2sim}
    \end{minipage}
    \hfill
     \begin{minipage}[b]{0.35\textwidth}
        \centering
        \includegraphics[width=\linewidth]{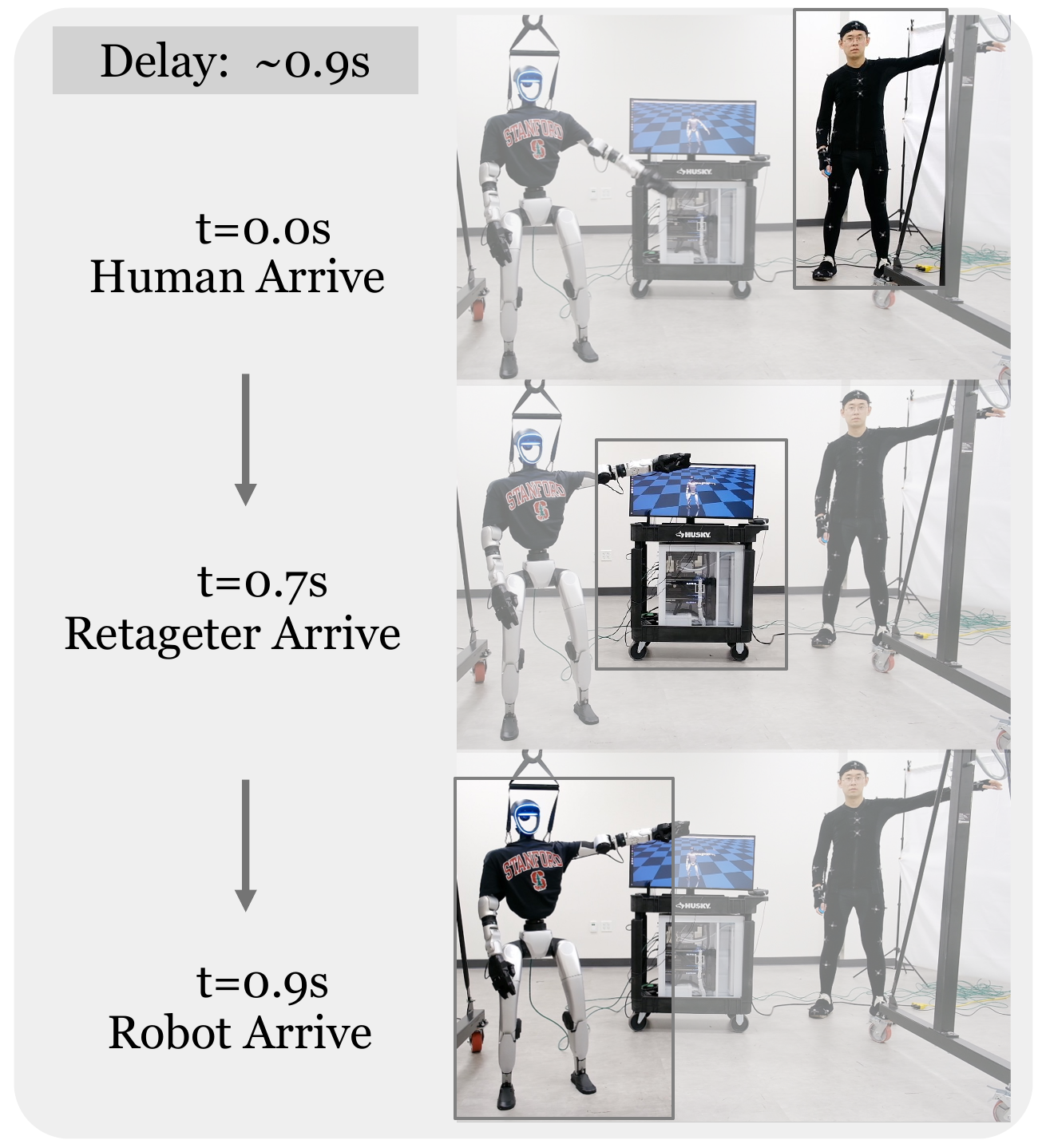}
        \caption{\textbf{Teleoperation delay} is roughly measured by the video, around 0.9 seconds.}
        \label{fig:teleop_delay}
    \end{minipage}
    \vspace{-0.2in}
\end{figure}

\section{Experiments}
In this section, we demonstrate the teleoperation capabilities of the \oursfull on Unitree G1~\citep{unitree_g1}, and analyze key factors affecting \ours's performance. We also show that \ours can transfer to other humanoid robots such as Booster T1~\citep{booster_t1_2025}. Additionally, we analyze \ours's reachability, teleoperation delay, and failure cases.

\subsection{Main Results on Whole-Body Humanoid Teleoperation}

We showcase the teleoperation capabilities of \ours on Unitree G1, a 29 DoF 1.3m humanoid robot. As shown in Figure~\ref{fig:task illustration} and \href{https:humanoid-teleop.github.io}{our website}, 
our robot can be teleoperated to perform diverse human-like whole-body skills, including:
\textbf{whole-body manipulation}, such as uprighting a fallen trash can, crouching to lift a box from the ground, and carrying a "Minions" toy from a table to a human;
\textbf{legged manipulation}, such as kicking a door shut/open while carrying objects with hands, delivering a powerful kick to a soccer ball, and transporting the box with its feet;
\textbf{locomotion}, such as executing sidesteps and backward walking, and crouching to navigate under obstacles;
\textbf{expressive motion}, such as boxing and performing Waltz dance steps with a human partner.

\begin{figure}[tb]
    \centering
   \includegraphics[width=1.0\linewidth]{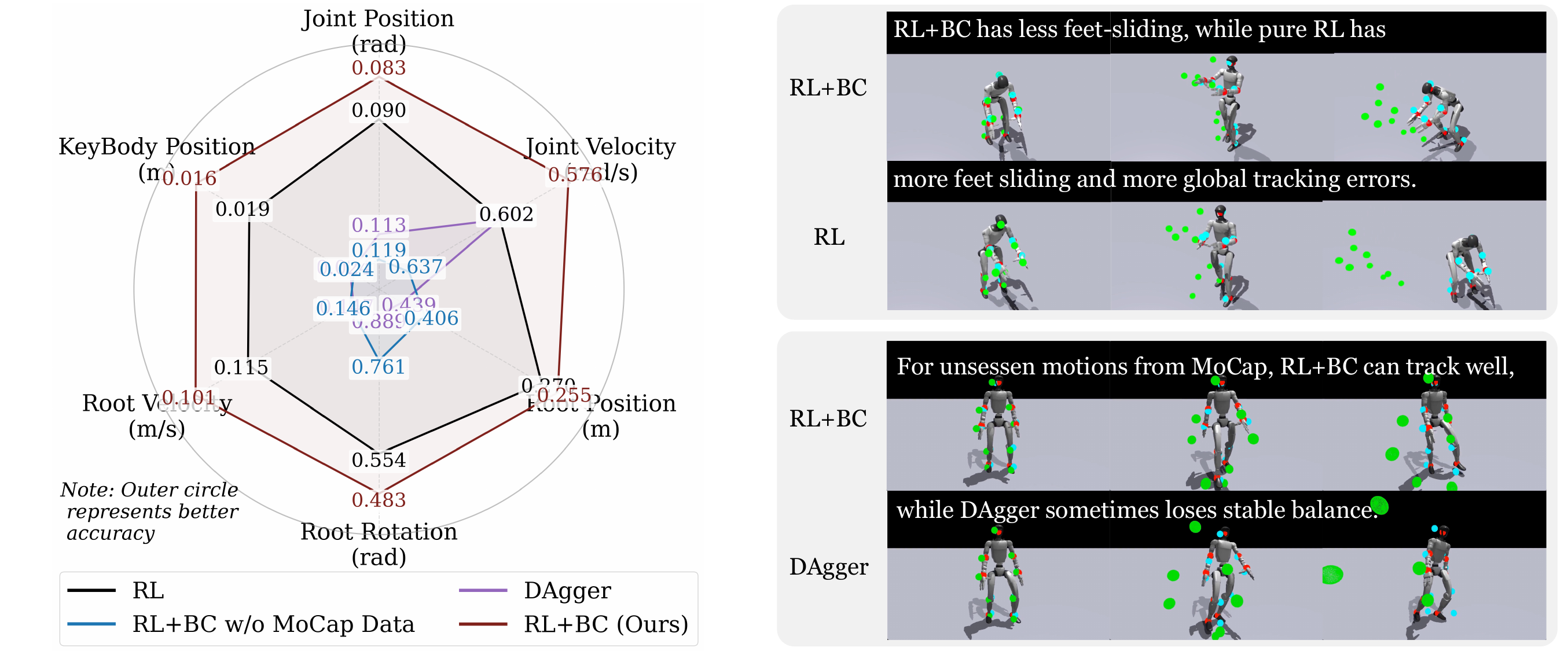}
    \vspace{-0.15in}
    \caption{(left) \textbf{Tracking errors of different controllers,} measured on our MoCap test data. (right) \textbf{Controller behaviors.} Our RL+BC controller produces smooth and robust behaviors.}
    \label{fig:rl+bc comparison}
    \vspace{-0.1in}
\end{figure}

To demonstrate that \ours functions as a general framework for diverse embodiments, we further evaluate it on Booster T1~\citep{booster_t1_2025}. Figure~\ref{fig:booster sim2sim} displays the sim-to-sim evaluation results. The controller successfully tracks diverse motions, including arm swinging with coordinated whole-body joints, deeply crouching down, and walking.

\begin{figure}[tb]
    \centering
    \includegraphics[width=1.0\linewidth]{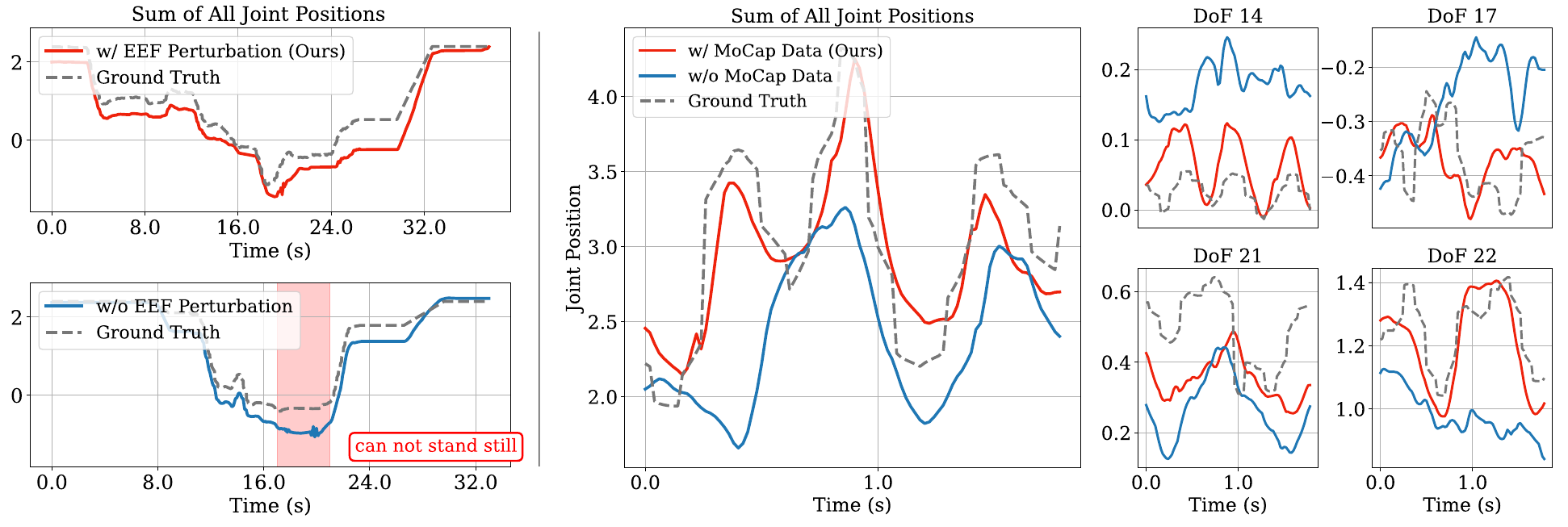}
    \vspace{-0.1in}
    \caption{(left) \textbf{Rollout curves} in the real world when the robot holds a box. (right) \textbf{Rollout curves} in MuJuCo from different controllers  when tracking MoCap data.}
    \label{fig:rollout curve}
    \vspace{-0.1in}
\end{figure}

\subsection{Ablation Experiments}

To further evaluate the impact of different components of \ours on its effectiveness in teleoperation, we collect another small-scale MoCap motion dataset (50 clips) for evaluation, all of which are not used in training. The following experimental results are evaluated on this test dataset.

\textbf{Key Finding 1: RL+BC $>>$
 RL $>>$ BC (DAgger).} Unlike HumanPlus~\citep{fu2024humanplus}, which uses single-stage RL, and OmniH2O~\citep{he2024omnih2o}, which employs DAgger, we find that our combined pipeline, RL+BC, achieves superior tracking accuracy (Figure~\ref{fig:rl+bc comparison} (left)) and motion smoothness (Figure~\ref{fig:rl+bc comparison} (right)). Pure RL approaches frequently exhibit feet sliding artifacts due to their inability to anticipate future motion goals. Meanwhile, DAgger can not stably and robustly track unseen motions occasionally, due to the lack of task reward guidance like RL.
 In summary, while RL demonstrates better generalization than BC~\citep{chu2025sft}, the combination of both approaches yields significant performance improvements.

\textbf{Key Finding 2: In-House MoCap Data Matters.} 
We find that adding even a small set of in-house MoCap sequences—retargeted online to mimic real teleoperation—substantially reduces tracking errors on unseen motions (Figure~\ref{fig:rl+bc comparison} (left) and Figure~\ref{fig:rollout curve} (right)). This gain arises from two factors: (1) our in-house captures are inherently noisier and less stable, suffering from calibration drift and occlusions; and (2) our online retargeter yields less-smooth reference motions compared to the offline version. Exposing the controller to these real-world imperfections bridges the gap between clean public datasets and the variability encountered during teleoperation, improving generalization.

\textbf{Key Finding 3: Learning to Apply Force.} We find that training policies without end-effector perturbations leads to drift and instability during stationary poses (Figure~\ref{fig:rollout curve} (left)). Incorporating perturbations during training significantly improves stability, especially crucial for contact-rich tasks.

\textbf{Key Finding 4: Better Online Retargeter, Better Tracking.} As detailed in Section~\ref{sec:curatemotiondata}, improving our online retargeter by simultaneously optimizing 3D positions and orientations yields smoother humanoid reference motions. Figure~\ref{fig:bar chart} (left) shows that such smoother motions benefit controllers trained both with and without MoCap data to reduce overall tracking errors.

\subsection{System Analyses}
\label{sec:system analysis}

\begin{figure}[t]
    \centering
    \includegraphics[width=1.0\linewidth]{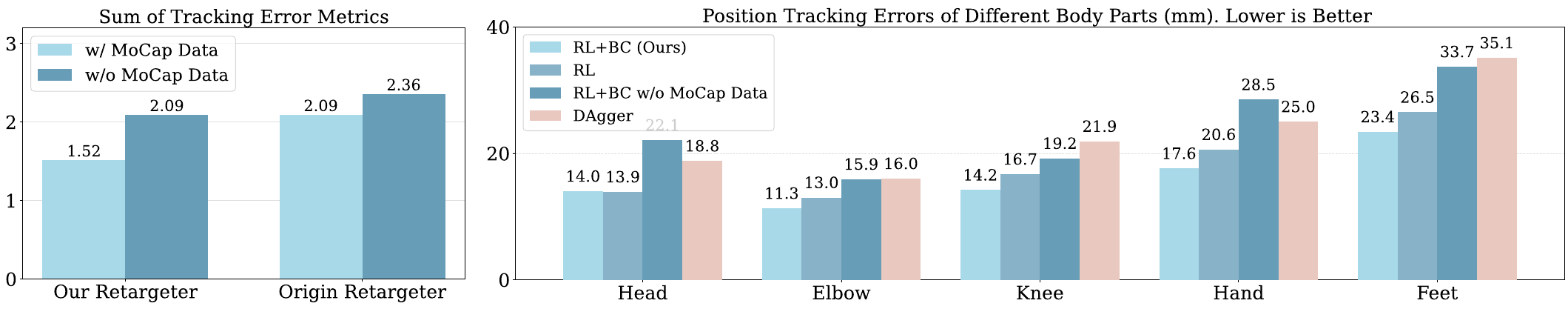}
    \vspace{-0.2in}
    \caption{(left) \textbf{Sum of tracking error metrics} on different online retargeters. (right) \textbf{Tracking errors across different body parts.} The feet exhibit the largest errors, highlighting the greater difficulty in accurately tracking lower-body movements.}
    \label{fig:bar chart}
    \vspace{-0.15in} 
\end{figure}
\textbf{Tracking Error Distribution Across Body Parts.} We are curious about how the controller's performance varies in tracking different body parts. As depicted in Figure~\ref{fig:bar chart} (right), we observe: (1) the end-effectors—hands and feet—exhibit the largest tracking errors, consistent with their positions at the extremities of the humanoid's kinematic tree; and (2) lower-body parts, such as feet and knees, generally incur higher tracking errors compared to upper-body parts like elbows and hands, confirming our intuition that tracking the lower body is inherently more challenging due to more complex contact dynamics.

\textbf{Teleoperation Delay.} The total teleoperation delay of our system is approximately 0.9 seconds, as measured in Figure~\ref{fig:teleop_delay}. The major overhead arises from generating tracking goals (0.7 seconds), while policy inference remains efficient (0.2 seconds). Reducing this latency further will be a key focus in future improvements.

\begin{figure}[htbp]
    \centering
\includegraphics[width=1.\linewidth]{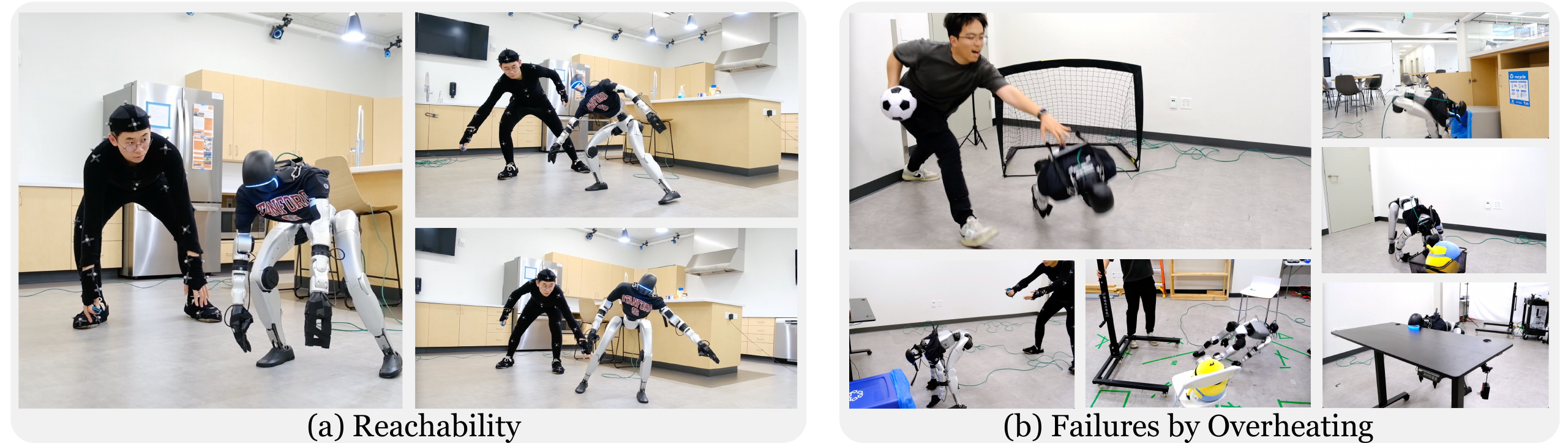}
    \vspace{-0.15in}
    \caption{\textbf{(a) Extreme reachability} by \ours. \textbf{(b) Failures} caused by motor overheating.}
    \label{fig:reach and fail}
    \vspace{-0.1in}
\end{figure}

\textbf{Reachability.} By utilizing whole-body DoFs, our teleoperation system significantly enhances reachability compared to prior work~\citep{he2024h2o,fu2024humanplus,he2024omnih2o}.
As illustrated in Figure~\ref{fig:reach and fail} (a), the robot can nearly reach its toes with its hands, demonstrating exceptional whole-body reachability.

\textbf{Failure Cases.} Most failures arise from hardware unreliability. In particular, our robot's motors tend to overheat after 5-10 minutes of continuous operation, especially during tasks that require crouching (Figure~\ref{fig:reach and fail} (b)), which necessitates cooling periods between tests. Nevertheless, our policy exhibits robust balance control even under extreme testing conditions.

\section{Conclusion}
In this work, we introduce the \oursfull, a system that teleoperates real-world humanoid robots using whole-body human motion data captured from MoCap devices. We provide systematic analysis of our controller training pipeline and overall system to identify critical factors for \ours. Our system fully leverages the robot's whole-body joints, enabling the execution of versatile and coordinated whole-body skills previously unachieved in the literature. The system demonstrates exceptional reachability compared to prior work, significantly expanding the capabilities of humanoid robots.
Our future work will include studying how to learn visuomotor policies with combined \ours-collected teleoperation data and egocentric human data.

\section{Limitations}

While \ours demonstrates strong capabilities in teleoperating humanoid robots, there remain several limitations and directions for future improvement:

\textbf{Lack of Robotic Feedback.} Currently, there is no robotic egocentric vision streamed back to the human operator. As a result, when visual occlusion occurs, it becomes challenging to teleoperate effectively. In addition, there is no tactile feedback to inform the operator when grasping actions succeed, limiting the naturalness and reliability of manipulation.

\textbf{Hardware Reliability.} As discussed in Section~\ref{sec:system analysis}, the current generation of humanoid hardware cannot sustain long-term continuous operation. Future improvements in robot hardware are likely to significantly enhance the overall system capability and robustness.

\textbf{Dependence on MoCap Systems.} Our method relies on a motion capture system, which is not portable and difficult to democratize, despite showcasing the potential of using whole-body human motion data. In the future, we plan to explore RGB-based pose estimation techniques to bridge the gap and approximate the quality of MoCap data more accessibly.
\acknowledgments{
We would like to thank all members of the CogAI group and The Movement Lab from Stanford University for their support, Sirui (Eric) Chen for his help with the video shooting and real-world experiments, and Haoyu Xiong for his helpful discussions. We also thank Stanford Robotics Center for providing the experiment space and the MoCap devices.
This work is in part supported by ONR MURI N00014-24-1-2748, NSF:FRR 2153854, Stanford HAI, and Stanford Wu-Tsai Human Performance Alliance.
}
\bibliography{main}

\end{document}